\documentclass{Interspeech}
\usepackage{amsmath,graphicx}

\usepackage{cite}
\usepackage{amssymb}
\usepackage{amsfonts}

\usepackage{multirow}
\usepackage{makecell}
\usepackage{color}
\usepackage{pifont}
\usepackage{comment}
\usepackage{booktabs}
\usepackage{soul}
\usepackage{cancel}

\newcommand{\cmark}{\ding{51}}%
\newcommand{\xmark}{\ding{55}}%

\usepackage{todonotes}
\newcommand{\tim}[1]{{\textcolor{blue}{#1}}}

\interspeechcameraready

\title{Exploring SSL Discrete Speech Features for Zipformer-based Contextual ASR}

\author[affiliation={1}]{Mingyu}{Cui}
\author[affiliation={2}]{Yifan}{Yang}
\author[affiliation={1}]{Jiajun}{Deng}
\author[affiliation={1}]{Jiawen}{Kang}
\author[affiliation={1}]{Shujie}{Hu}
\author[affiliation={1}]{Tianzi}{Wang}
\author[affiliation={1}]{Zhaoqing}{Li}
\author[affiliation={3}]{Shiliang}{Zhang}
\author[affiliation={2}]{Xie}{Chen}
\author[affiliation={1}]{Xunying}{Liu}

\affiliation{}{The Chinese University of Hong Kong}{Hong Kong SAR, China}
\affiliation{}{X-LANCE Lab, School of Computer Science, Shanghai Jiao Tong University}{China}
\affiliation{}{Alibaba}{China}
\email{\{mycui,xyliu\}@se.cuhk.edu.hk}

\usepackage{comment}

\begin{document}

\maketitle

\begin{abstract}
This paper investigates discrete tokens based cross-utterance speech contexts modelling for Zipformer-Transducer (Z-T) systems. Their efficacy and efficiency in modelling preceding, current and future speech utterance contexts using concatenation or pooling projection of Z-T encoder embeddings are extensively shown on the 1000-hr GigaSpeech-M and DementiaBank Pitt elderly speech datasets over comparable contextual Z-T baselines using filterbank or continuous WavLM features. The best-performing discrete tokens based contextual Z-T system outperforms the non-contextual baseline by statistically significant average WER reductions of 0.39\% and 1.41\% absolute (3.4\% and 3.4\% relative) on the two tasks, respectively.  Model training time speedup ratios up to 4.36x is obtained over continuous WavLM feature-based  contextual Z-T systems, while retaining up to 98.0\% of their WER reductions over non-contextual baselines\footnote{\small{Our work is publicly available at https://github.com/open-creator/icefall/tree/master/egs/gigaspeech/Context\_ASR.}}.
\end{abstract}
%

\section{Introduction}
\label{sec:intro}
Context plays an important role in natural speech processing tasks including ASR. The incorporation of long-range cross-utterance speech contexts in ASR systems, in addition to speech contained within the current utterance being processed, has been widely shown to improve ASR 
performance \cite{wei2022leveraging, kim2018dialog, hou2022bring, chang2023context, li2022recent, chen2021developing}. Among the existing works, the benefits of incorporating cross-utterance textual contexts have been reported in language modelling tasks \cite{irie2019training, xiong2018session, dai2019transformer, liu2020contextualizing, liu2013use, beltagy2020longformer,sun2021transformer}, the predictor module of neural Transducers fusing BERT embeddings \cite{chang2021context}, and the decoder module of factorized Transducers \cite{chen2022factorized, gong2023longfnt, gong2024advanced} using RoBERTa text embeddings. 

Discrete token-based speech features provide compact representations. They have been successfully applied to multiple applications such as automatic speech recognition (ASR) \cite{wang2025phonepurityguideddiscretetokens} and text-to-speech (TTS)\cite{baevski2019effectiveness, chang2023exploration, yang2023towards, guo2025recent, wang2024evaluating, shen2024acoustic, lee2024high}. 
Discrete token features have been extensively used in non-contextual ASR systems \cite{wang2025phonepurityguideddiscretetokens, chang2024interspeech2024challengespeech, chang2023exploringspeechrecognitiontranslation, chen2024loss}, while their potential in learning longer-range, cross-utterance speech contexts remains underexplored.




To this end, discrete tokens extracted from fine-tuned WavLM \cite{chen2022wavlm} self-supervised learning (SSL) models are used as cross-utterance speech contexts features in the Encoder of Zipformer-Transducer (Z-T) ASR systems. 
Their efficacy and efficiency in modelling preceding, current, and future speech utterance contexts using either concatenation or pooling projection of Z-T encoder embeddings are extensively shown on the 1000-hr GigaSpeech-M and DementiaBank Pitt elderly speech datasets over comparable contextual Z-T baselines using filterbank (FBank) or continuous WavLM features. The best-performing discrete token-based contextual Z-T system outperforms the non-contextual baseline by statistically significant average WER reductions of \textbf{0.39}\% and \textbf{1.41}\% absolute (\textbf{3.4}\% and \textbf{3.4}\% relative) on the two tasks, respectively.  Model training time speedup ratios up to \textbf{4.36}x are obtained over continuous WavLM feature-based contextual Z-T systems, while retaining up to \textbf{98.0}\% of their WER reductions over the comparable non-contextual baselines.

The main contributions of this paper are as follows: 
\textbf{1)}	To the best of our knowledge, this work pioneers the use of SSL pre-trained discrete speech features for modelling preceding, current, and future speech utterance contexts in Z-T ASR systems. In contrast, the use of SSL pre-trained speech features in prior research has been limited to modelling current utterance contexts only for ASR and TTS tasks \cite{baevski2019effectiveness, chang2023exploration, yang2023towards}. 
\textbf{2)} Systematic investigations are conducted on benchmark ASR tasks across two domains using the 1000-hr GigaSpeech-M and the low-resource DementiaBant Pitt elderly speech corpora, demonstrating the efficacy and efficiency of using SSL discrete tokens for context modeling through side-by-side comparisons of input features, context fusion methods and operating positions.

\begin{figure*}
    \centering
    \includegraphics[width=5.3in]{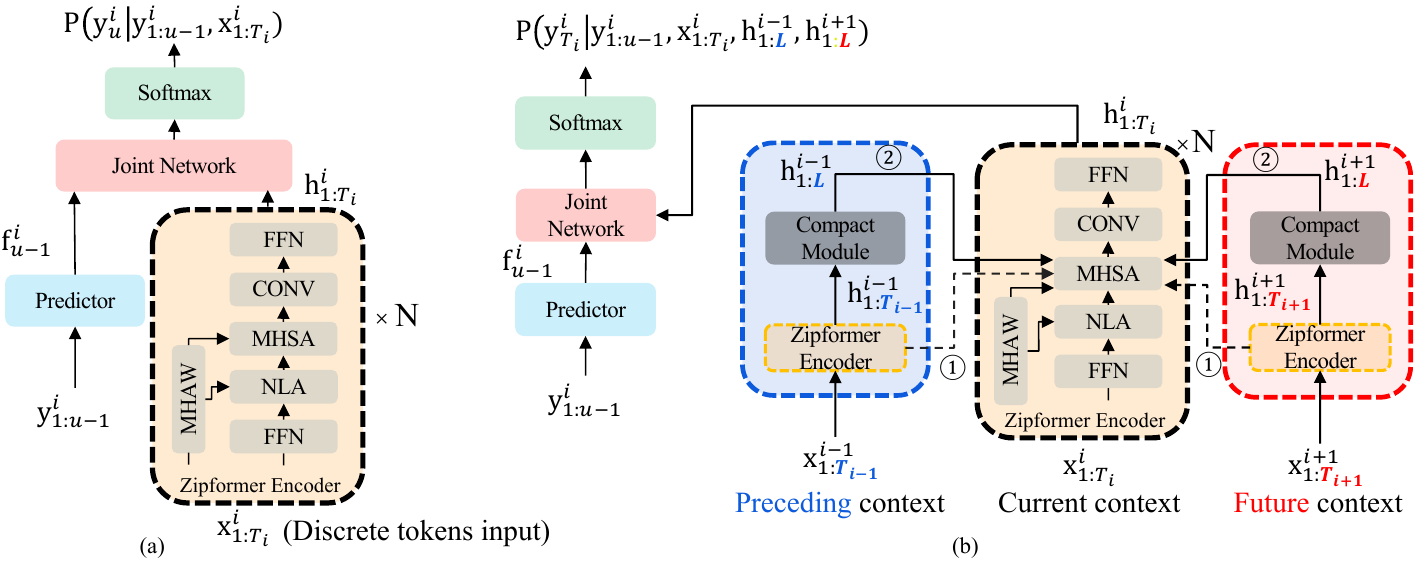}
    \caption{Examples of: a) Standard Zipformer-Transducer models using current utterance context only with discrete tokens as input in Sec. 2; b) Zipformer-Transducer using cross-utterance speech contexts of the most recent \textcolor{blue}{preceding} $(i-1)^{\rm th}$ utterance $\mathbf{x}_{1:T_{i-1}}^{i-1}$ of $T_{i-1}$ frames (blue dotted box) and \textcolor{red}{future} $(i+1)^{\rm th}$ utterance $\mathbf{x}_{1:T_{i+1}}^{i+1}$ of $T_{i+1}$ frames (red dotted box) in the Zipformer Encoder. The black dotted line \ding{172} denotes the concatenation of cross-utterance Encoder embeddings described in Sec. 4.1, while the black line \ding{173} via the "Compact Module" denotes cross-utterance Encoder embeddings pooling projection described in Sec. 4.2.}
    
    \label{fig:all}
\vspace{-0.5cm}
\end{figure*}
\vspace{-0.2cm}
\section{Zipformer-Transducer Based ASR }

This paper utilizes the neural Transducer \cite{graves2012sequence} model to perform speech recognition, which is composed of three modules: audio ``Encoder", text ``Predictor" and ``Joint Network" respectively, as depicted in Fig. \ref{fig:all}(a).  Here we denote  $\mathbf{x}_{1:T_i}^{i}=[\mathbf{x}_{1}^{i}, \mathbf{x}_{2}^{i},\cdots, \mathbf{x}_{T_i}^{i}]$ as the $i$-th utterance of an audio clip or conversation session with $T_i$-frames and $\mathbf{y}_{1:U_i}^{i}=[\mathbf{y}_{1}^{i}, \mathbf{y}_{2}^{i},\cdots,\mathbf{y}_{U_i}^{i}]$ as the corresponding label of length $U_i$. SSL pre-trained discrete token sequence $\mathbf{{x}}_{1:T_i}^{i} = [x^i_1,x^i_2,\cdots, x^i_{T_i}]$ of length $T_i$ is fed into the Encoder to produce the acoustic representation $\mathbf{{h}}_{1:{T_i}}^{i}$. Note that each element in the discrete token sequence $\mathbf{{x}}_{1:T_i}^{i}$ is an integer (codebook index) rather than a vector. The history output labels $\mathbf{{y}}_{1:u-1}^{i}$ are fed into the predictor module to generate the text representation $\mathbf{f}_{u-1}^{i}$. The outputs of the Encoder and predictor are then combined in the Joint Network via a non-linear function such as ReLU to obtain the hidden state $\mathbf{g}_{t, u-1}^{i}$ at time step $t$ with output history $\mathbf{y}_{1:u-1}^{i}$. These operations are as follows,
\vspace{-0.3cm}
\begin{equation}
    \begin{aligned}
        \mathbf{h}_{1:T_i}^{i} &= \mathrm{Encoder}(\mathbf{x}_{1:{T_i}}^{i}) \\ 
        \mathbf{f}_{u-1}^{i} &= \mathrm{Predictor}(\mathbf{y}_{1:u-1}^{i}) \\
        \mathbf{g}_{t,u-1}^{i} &= \mathrm{Relu}(\mathbf{h}_{1:T_i}^{i} + \mathbf{f}_{u-1}^{i}) \\
        P(\mathbf{y}_{t}^{i}| \mathbf{y}_{1:u-1}^{i}, \mathbf{x}_{1:{T_i}}^{i}) &= \mathrm{Softmax}({\mathbf W}_{o} * \mathbf{g}_{t,u-1}^{i})      
    \end{aligned}
\end{equation}
where $\mathbf{W}_{o}$ is a linear transformation applied prior to the final Softmax output layer. In this paper, the Z-T architecture using Zipformer based Encoder and Stateless \cite{ghodsi2020rnn} Prediction modules are used throughout this paper.


\section{SSL Discrete Speech Features}
In this work, we employ the WavLM model and k-means method for speech discretization. Specifically, we use the WavLM-Large model and extract hidden embeddings from the final Transformer Encoder layer based on the semantic correlation analysis in \cite{chen2022wavlm}. We employ 2000 clusters for the k-means clustering, as previous findings \cite{chang2023exploration} suggest that using more discrete tokens can improve ASR performance and phone-normalized mutual information (PNMI).

We employ discrete speech tokens of each type with their corresponding texts to train the Z-T model by leveraging the Recurrent Neural Network Transducer (RNN-T) loss. A linear embedding layer projects these discrete token sequences to 80 dimensions. These features are then fed into the ASR model to perform the training process. Furthermore, we perform data augmentation using four types of perturbation operations: Time Warping, Time Masking, Embedding Masking, and Gaussian Noise. More 
details can be found in \cite{yang2023towards}.

\section{Discrete Tokens for Contextual Z-T}
In this section, we propose cross-utterance speech contexts conditioned  Z-T models with preceding and future contextual representations.
%

\subsection{Concatenation of Encoder Embeddings }
\label{cross-utterance-embeddings}
A common practice is to utilize the complete outputs of Conformer Encoder obtained from the preceding utterance(s) \cite{cui23_interspeech}. These are then concatenated and serve as the long span context representation to augment the current utterance's input features before applying the linear transformations to produce the query, key, and value vectors. After incorporating the preceding utterance(s)' contexts in the MHSA module, the current utterance context-based Z-T model is modified as follows during training and evaluation:
\begin{equation}
    \begin{aligned}
        \mathbf{h}^{i-1}_{1:T_{i-1}} &= \mathrm{Encoder}(\mathbf{x}^{i-1}_{1:T_{i-1}}) \\
        \hat{\mathbf{x}}^{i}_{1:T_i} &= \mathrm{FFN}(\mathbf{x}^{i}_{1:T_i}) \\
        \mathbf{h}^{i}_{1:T_i} &= \mathrm{MHSA}(\hat{\mathbf{x}}^{i}_{1:T_i}, \mathbf{h}^{i-1}_{1:T_{i-1}}) \\
    \end{aligned}
    \label{Encoder}
\end{equation}
We stacked the inputs before each corresponding MHSA layer in the Encoder to carry over contextual information. This process (Eq. 2, Line 3) is consistently applied across all layers, even though the detailed operations for each layer are not explicitly depicted in the formula. An example of Z-T models using such complete preceding utterances’ Encoder contextual features is shown in Fig. \ref{fig:all}(b) (blue dotted box, via black dotted line marked with \ding{172}). 

\subsection{Pooling Projection of Encoder Embeddings }
\label{cross-utterance-pooling}
Specially designed attention pooling layers ~\cite{cui23_interspeech} are applied over preceding utterances' Encoder contextual vectors to project the complete cross-utterance context ( as described in Sec. 4.1 ) into \tim{a} partial cross-utterance speech contexts. The previous complete utterances' Encoder hidden context vectors are cached prior to the attention-based pooling project operations. Let the Zipformer Encoder's outputs be $\mathbf{h}^{i-1}_{1:{T_{i-1}}} \in \mathbb{R}^{T_{i-1} \times D}$ for the preceding $(i-1)^{\rm th}$ utterance of $T_{i-1}$ frames, where ${D}$ stands for the Encoder output vector dimensionality. The cross-utterance Encoder contextual states are attention-pooled and projected to low-dimensional partial representations as:
\begin{equation}
    \begin{aligned}
        \mathbf{h}^{i-1}_{1:T_{i-1}} &= \mathrm{Encoder}(\mathbf{x}^{i-1}_{1:T_{i-1}}) \\
        \mathbf{h}^{i-1}_{1:L} &= \mathrm{Compact\ Module}(\mathbf{h}^{i-1}_{1:T_{i-1}})
    \end{aligned}
\end{equation}
The resulting compact, fixed-length $L \times D$ cross-utterance Encoder contextual features are combined with the current utterance in the Zipformer MHSA module. An example of Z-T models using the compressed cross-utterance contexts is shown in Fig.~\ref{fig:all}(b) (blue dotted box, via black line marked with \ding{173}). 

\subsection{Concatenation of Future Context Embeddings}

In order to learn richer contextual information, we propose to utilize future cross-utterance speech contexts embeddings in addition to the preceding cross-utterance speech contexts. These future context embeddings are being concatenated with the preceding context embeddings to augment the training performance, following similar approaches as described in Sec. \ref{cross-utterance-embeddings} and Sec. \ref{cross-utterance-pooling}. To incorporate both preceding and future cross-utterance speech contexts, we modify the current utterance context-based Z-T model as follows:
\begin{equation}
    \begin{aligned}
        \mathbf{h}^{i-1}_{1:T_{i-1}} &= \mathrm{Encoder}(\mathbf{x}^{i-1}_{1:T_{i-1}}) \\
        \mathbf{h}^{i+1}_{1:T_{i+1}} &= \mathrm{Encoder}(\mathbf{x}^{i+1}_{1:T_{i+1}}) \\
        \hat{\mathbf{x}}^{i}_{1:T_i} &= \mathrm{FFN}(\mathbf{x}^{i}_{1:T_i}) \\
        \mathbf{h}^{i}_{1:T_i} &= \mathrm{MHSA}(\hat{\mathbf{x}}^{i}_{1:T_i}, \mathbf{h}^{i-1}_{1:T_{i-1}}, \mathbf{h}^{i+1}_{1:T_{i+1}}) \\
    \end{aligned}
    \label{Encoder}
\end{equation}
An example of Z-T using future context embeddings is shown in ~Fig.~\ref{fig:all}(b) (red dotted box, via black dotted line marked with \ding{172} or black line marked with \ding{173}).

\vspace{-0.2cm}
\section{Experiments}
\vspace{-0.1cm}
\subsection{Experiment Setups}
The GigaSpeech M-size corpus \cite{chen2021gigaspeech} with 1000-hr speech collected from Audiobook, Podcast, and YouTube is used for pre-training. The standard dev and test sets contain 12 and 40 hours of audio, respectively. The training set of the DementiaBank Pitt \cite{becker1994natural} corpus consists of 27.2 hours of recorded interviews between 
292 elderly participants (Par) and the clinical investigators (Inv)
after silence stripping, and is further expanded to 59 hours via speed perturbation for fine-tuning~\cite{ye2021development}. The development and test sets contain 2.5 hours and 0.6 hours of audio, respectively.
The training objective is the pruned RNN-T loss. A 6-stack Zipformer \cite{yao2023zipformer} is used by the Encoder, with downsampling factors of (1,2,4,8,4,2). A stateless predictor, comprising an embedding layer and a 512-dim Conv1D layer, is utilized as the label predictor. A convolution subsampling module with a stride of 2 is placed to reduce
the frame rate to 50 Hz before the input feature is fed into the Encoder. The model comprises 65.5M parameters in total. All models are trained on 8 x NVIDIA H20 96GB GPUs. 

In FBank-based experiments, SpecAugment is applied for robustness. The input is 80-channel FBank features extracted over windows of 25ms strided by 10ms. 500 byte-pair-encoding (BPE) tokens were used for pre-training and fine-tuning datasets. In continuous WavLM feature-based experiments, we directly utilized the continuous hidden representation extracted from the 21st layer of WavLM as input features. A linear projection layer is then applied to transform the continuous WavLM features to match the input dimension of the Zipformer ASR system. We also choose the 21st layer of WavLM for extracting discrete tokens. The selection of the 21st layer from the WavLM model is inspired by canonical correlation analysis (CCA) in \cite{pasad2023comparativelayerwiseanalysisselfsupervised}. In fine-tuning on elderly speech experiments, the parameters in the pre-trained Zipformer Encoder, Predictor, and Joint network are inherited and fine-tuned for the elderly speech dataset. A new Linear projection layer is reinitialized after the Joint network. To capture cross-utterance speech contexts, we serialized the training data of the same audio clip or conversation based on the utterances' start times. Significance tests are performed using the standard NIST implemented \cite{pallet1990tools} Matched Pairs Sentence-Segment Word Error (MAPSSWE) Test proposed by Gillick \cite{gillick1989some} with a significant level of $\alpha=0.05$ denoted by $\dagger, \triangle, \star$ throughout the experiments over the baselines. We set the dimension of the pooling projection (Eq. (3), bottom equation, Sec. \ref{cross-utterance-pooling}) as $L=32$, following the best setting in~\cite{cui23_interspeech} (Table 2, Sys.13 and 16), which offers a better WER-RTF trade-off compared to other settings, e.g., $L=8$ or $L=16$. The learning rate is set to 0.0002. We do not use any language models in our experiments.

\begin{table}[htbp]

 \caption{Performance contrasts between FBank features, discrete tokens features obtained from WavLM (Disc.), and continuous features derived from WavLM (Con.) on different concatenation positions. ``$\dagger$, $\star$, $\triangle$" denotes a statistically significant WER improvement over the corresponding baseline systems Sys.1, 2, 3 (determined by the type of  ``current utterance feature").}
\label{pretraining}
\centering
\scalebox{0.51}{
\renewcommand\arraystretch{1.5}
\begin{tabular}{ccccccccccc} 
\toprule
\multirow{2}{*}{\textbf{ID}} & 
\multirow{1}{*}{\textbf{Cntx.}} &
\multirow{1}{*}{\textbf{Prec. }} &
\multirow{1}{*}{\textbf{Current}}&
\multirow{1}{*}{\textbf{Future}} &
\multicolumn{3}{c}{\textbf{WER}} & 
\textbf{Training} &
\multicolumn{2}{c}{\textbf{Inference RTF}} \\
\cline{6-8}
& \textbf{Fusion}&\textbf{utt. feat.} &\textbf{utt. feat.} & \textbf{utt. feat.} &\textbf{Dev} &\textbf{Test} &\textbf{Avg.} &\textbf{time (\#hour)}& \textbf{SSL} & \textbf{ZFM } \\

\midrule

1   & - & - & FBank&- &  12.20 & 12.20 &12.20& 10.0&-&0.0278\\
2      & - & - &Con.  &- & 10.92 & 10.96 &10.94& 38.0& 0.29 & 0.0301 \\
3      & - & - &Disc. &- & 11.47 & 11.55 &11.53&6.7& 0.065 & 0.0280\\

\cline{2-11}
4      &  & FBank &FBank  &-&12.64 &12.66&12.65&11.5&-& 0.0280\\

5      & \multirow{2}{*}{Input} & Con. &Con. &-& 11.10& 11.13&11.12& 42.9&0.29&0.0296\\
6      & \multirow{2}{*}{Concat.} & Disc.&Disc. &-&11.75 & 11.80&11.78&7.9&0.065 &0.0282\\
7      & \multirow{2}{*}{\cite{hori2020transformer, hori2021advanced}}& FBank &FBank  &FBank & 12.70 & 12.73 &12.72&20.0 &- &0.0281\\
8      & & Con. &Con.  &Con. & 11.26 & 11.30&11.29&56.7&0.58&0.0299\\
9      & & Disc. &Disc.  &Disc. & 11.85&11.92&11.89&13.0&0.13&0.0283\\
\cline{2-11}
10      & \multirow{2}{*}{Cross-} & FBank &FBank  & \multirow{5}{*}{-} & $12.07^\dagger$ & $12.09^\dagger$&$12.08^\dagger$&10.4&-& 0.0279\\
11      & \multirow{2}{*}{utterance}& Disc. &FBank  &   &  $12.03^\dagger$ & $12.03^\dagger$&$12.03^\dagger$&10.2& 0.065& 0.0281  \\

12     & \multirow{2}{*}{pooling}& FBank &Disc.  &   &  11.40 & 11.51 &11.46&10.2&0.065& 0.0275\\
13      & \multirow{2}{*}{projection} & Con. &Con.  &&10.81 & 10.83&10.82& 40.0&0.29&0.0292\\
14      &  & Disc. &Disc.  &   & $11.32^\triangle$ & $11.40^\triangle$ & $11.37^\triangle$ &7.1&0.065& 0.0281\\
\cline{2-11}
15      & \multirow{2}{*}{Cross-} & FBank &FBank  &\multirow{5}{*}{-} & $12.01^\dagger$ & $12.00^\dagger$ &$12.00^\dagger$ &11.3&-& 0.0283\\
16      & \multirow{2}{*}{utterance}& Disc. &FBank  & & $11.95^\dagger$ & $11.98^\dagger$ & $11.97^\dagger$&11.0&0.065& 0.0279\\
17      & \multirow{2}{*}{Encoder} & FBank &Disc.  & & 11.37 & 11.49 &11.46 &11.0&0.065& 0.0284\\
18      & \multirow{2}{*}{embedding}& Con. &Con.  &   & $10.69^\star$ & $10.70^\star$&$10.70^\star$&41.7&0.29&0.0291\\
19      &  & Disc. &Disc.  &  & $\textbf{11.21}^\triangle$& $\textbf{11.30}^\triangle$ &$\textbf{11.28}^\triangle$ &8.3&0.065& 0.0285 \\
\cline{2-11}
20      & \multirow{2}{*}{Cross-} & FBank &FBank  & FBank & $12.03^\dagger$ & $12.03^\dagger$ & $12.03^\dagger$&18.3&-& 0.0290\\
21      & \multirow{2}{*}{utterance}& Disc. &FBank  & FBank  & $11.95^\dagger$ & $11.94^\dagger$ &$11.94^\dagger$ &17.6&0.065& 0.0286\\
22      & \multirow{2}{*}{pooling}& FBank &Disc.  &Disc.   & $11.33^\triangle$ & $11.40^\triangle$ &$11.38^\triangle$ &16.0&0.13& 0.0287\\
23      &  \multirow{2}{*}{projection}& Con. &Con.  & Con.  & $10.70^\star$ & $10.71^\star$& $10.71^\star$ &53.3&0.58&0.0305\\
24      &  & Disc. &Disc.  & Disc.  & $11.25^\triangle$ & $11.28^\triangle$ & $11.27^\triangle$ &12.3&0.13& 0.0294\\
\cline{2-11}
25      & \multirow{2}{*}{Cross-} & FBank &FBank  &FBank & $11.96^\dagger$ & $11.95^\dagger$&$11.95^\dagger$&20.7&-& 0.0288\\
26      & \multirow{2}{*}{utterance}& Disc. &FBank  &FBank & $11.85^\dagger$ & $11.90^\dagger$&$11.89^\dagger$&20.0&0.065 & 0.0290   \\
27      & \multirow{2}{*}{Encoder} & FBank &Disc.  &Disc. & $11.30^\triangle$ & $11.32^\triangle$ &$11.31^\triangle$ &18.6&0.13&0.0287 \\
28      & \multirow{2}{*}{embedding}& Con. &Con.  & Con.  & $10.53^\star$ & $10.54^\star$&$10.54^\star$&55.0&0.58&0.0302\\
29      &  & Disc. &Disc.  & Disc. & $\textbf{11.15}^\triangle$ & $\textbf{11.14}^\triangle$ &$\textbf{11.14}^\triangle$ &12.6&0.13 & 0.0292\\

\bottomrule
\end{tabular}}
\vspace{-0.5cm}
\end{table}

\subsection{Performance on the GigaSpeech 1000h Dataset}
Table~\ref{pretraining} presents the performance of baseline and various contextual Z-T models on the GigaSpeech M-size corpus 
using FBank, WavLM continuous or discrete token based input features. Several trends can be observed:
\textbf{1)} Incorporating preceding, current, and future speech utterance contexts using concatenation or pooling projection of Z-T encoder embeddings consistently brings statistically significant performance improvements over the non-contextual baseline, regardless of the concatenation position (preceding or future segments) or the input feature type (FBank, continuous features, or discrete tokens) when comparing Sys.10-29 (using context concatenation) with Sys.1-3 (using current utterance context only).

\textbf{2)} Systems that concatenate context embeddings from preceding and future utterances (Sys.20-29) achieve better performance than systems that use fewer concatenation positions, such as those utilizing only preceding contexts (Sys.10-19). This trend indicates that incorporating context information from both preceding and future contexts is more effective than considering only the preceding context. In particular, the cross-utterance Encoder embedding concatenation approach, utilizing WavLM discrete token features for preceding, current, and future utterances' contexts (Sys.29), achieves a SOTA performance. On the dev and test sets, it outperforms the comparable baseline Z-T system solely utilizing the current utterance context (Sys.3) by a statistically significant WER reduction of \textbf{0.32}\% and \textbf{0.41}\% absolute (\textbf{2.8}\% and \textbf{3.5}\% relative).


\textbf{3)} Systems using WavLM discrete token features as input (Sys.14, 19, 24, 29) outperform those with Fbank features (Sys.10, 15, 20, 25) when the concatenation positions are the same. This suggests that the discrete token features generated by the WavLM SSL model can capture richer semantic information, leading to better performance in cross-utterance speech contexts modelling compared to Fbank features as input.

\textbf{4)} Systems using WavLM discrete token features as input (Sys.14, 19, 24, 29) can obtain comparable WER reductions to the systems utilizing WavLM continuous features as input (Sys.13, 18, 23, 28). In particular, the cross-utterance Encoder embeddings approach, utilizing WavLM discrete token features for preceding, current, and future utterances' contexts (Sys.29), achieves model training time speedup ratios up to \textbf{4.36}x over the continuous WavLM feature-based contextual Z-T systems (Sys.28), while retaining up to \textbf{98.0}\% of their WER reductions over non-contextual baselines (Sys.29 vs. Sys.28).

\textbf{5)} The latency analysis for cross-utterance context modeling covers two aspects: \textbf{a) SSL discrete token feature forwarding}: approximately 0.065 real time for systems using preceding utterance contexts only, but not those of the future utterances (Sys.11,12,14,16,17,19), assuming preceding utterances discrete tokens are cached while processing the current utterance; approximately 0.13 real time when using both preceding and future utterance contexts (Sys.22,24,27,29), which is approximately \textbf{4.5}x faster than the SSL continuous feature based Z-T systems (Sys.23,28). \textbf{b) Inference RTF in Z-T}: our cross-utterance speech context-based systems (Sys.10-29) obtain an average RTF of 0.028, comparable to the non-contextual baseline systems (Sys.1-3) with current utterance context only \footnote{We attempted to implement MERL's method (Sys.4-9) for input context feature fusion, but the authors confirmed that their implementation is proprietary and not publicly available.}. 

\vspace{-0.3cm}
\begin{table}[htbp]
  \caption{Performance (WER\%) contrasts between the best cross-utterance speech context-based system and other SOTA results on the GigaSpeech-M (GS.) dataset.}
  \label{benchmark}
  \centering
  \scalebox{0.56}{
\renewcommand\arraystretch{1.3}
\begin{tabular}{cccccccc} 
\toprule
\multirow{2}{*}{\textbf{ID}} &

\multirow{2}{*}{\textbf{System}} &
\multicolumn{3}{c}{\textbf{Cntx.}}&
\textbf{Code} &
\multirow{2}{*}{\textbf{\#Parm.}} &
\textbf{GS.} \\
 &&\textbf{Fusion}&\textbf{Prec.}&\textbf{Future}&
 \textbf{Source}& &\textbf{Avg.}   \\
\midrule
 1&Kaldi-2021 TDNN \cite{chen2021gigaspeech} &\multirow{5}{*}{\xmark}&\multirow{5}{*}{\xmark}&\multirow{5}{*}{\xmark} &NA&NA& 17.75 \\

2 & CFM-AED-2024 \cite{fox2024updated} & & & &NA&NA&15.32 \\
 3&Zipformer-2023-Hubert \cite{yang2023towards} &&&&\ding{172}&65.7M& 14.62  \\

 4&Zipformer-2023 (TAB.1 Sys.1)\cite{yang2023towards}&&&&\ding{172}&65.7M& 12.20\\

 5&Zipformer-2023(TAB.1 Sys.3)&&&&\ding{173}&65.7M& 11.53 \\

\hline

 6&LongFNT-2023\cite{gong2023longfnt}&External Enc.&\multirow{4}{*}{\cmark}&\xmark&NA&NA& 14.60\\ 

 7&ESPnet-2023 CFM-Transducer\cite{cui23_interspeech} &Embed. Concat&&\xmark&NA&88.3M& 14.25 \\
 8& Zipformer-Cntx. (TAB.1 Sys.19)&Embed. Concat&&\xmark&\ding{173}&65.7M& \textbf{11.28} \\

 9&Zipformer-Cntx. (TAB.1 Sys.29)&Embed. Concat&&\cmark& \ding{173}&65.7M &\textbf{11.14} \\
\bottomrule
\end{tabular}}
\vspace{-0.5cm}
\end{table}

\subsection{Performance Benchmarking Against SOTA}
\vspace{-0.1cm}
The performance of the best SSL discrete token-based cross-utterance speech context system (Table~\ref{pretraining}, Sys.29) is further contrasted in Table~\ref{benchmark} with the state-of-the-art (SOTA) performance on the same tasks using the most recent hybrid and E2E systems reported in the literature to demonstrate their competitiveness. Sys.3-5, and Sys.8-9 are trained from scratch. The implementations for Sys.3-4 are derived from code source \ding{172}\footnote{https://github.com/k2-fsa/icefall/tree/master/egs/gigaspeech/ASR}. Our implementation of Sys.5, 8-9 is released in code source \ding{173}\footnote{https://github.com/open-creator/icefall/tree/master/egs/gigaspeech/ \\ Context\_ASR}.

\begin{table}[h]
  \caption{Performance of models fine-tuned on the DementiaBank Pitt elderly speech dataset with cross-utterance speech contexts. ``$\dagger$, $\star$, $\triangle$" denotes a statistically significant WER improvement over the corresponding baseline (Sys.1, 2, 3).}
  \label{fine-tuning}
  \centering
  \scalebox{0.45}{
\renewcommand\arraystretch{1.3}
\begin{tabular}{cccccccccccc} 
\toprule
\multirow{3}{*}{\textbf{ID}} &
\multirow{2}{*}{\textbf{Context}} & 
\textbf{Prec.} &
\textbf{Cur.}  & 
\textbf{Future} & \multirow{2}{*}{\textbf{Fine-tuned}}&\multicolumn{5}{c}{\textbf{WER}} &\multirow{2}{*}{\textbf{Fine-tuning}}\\
\cline{7-11}
&\multirow{2}{*}{\textbf{Fusion}}& \textbf{ utt.}&\textbf{ utt.} &\textbf{ utt.} & \multirow{2}{*}{\textbf{from}} &\multicolumn{2}{c}{\textbf{DEV}} &\multicolumn{2}{c}{\textbf{TEST}} &\multirow{2}{*}{\textbf{Avg.}}&\multirow{2}{*}{\textbf{time (\#hour)}}\\
&&\textbf{feat.}&\textbf{ feat.}&\textbf{ feat.} &&\textbf{INV.}& \textbf{PAR.}& \textbf{INV.}&\textbf{PAR.}\\
\midrule

1 & - & - & Fbank & - & Tab. \ref{pretraining}.1 & 27.33&60.69&24.56 & 49.10 & 44.19&1.0\\

2 & - & - & Con. & - & Tab. \ref{pretraining}.2& 24.16 & 53.66 & 21.76 & 43.38& 39.54&4.5\\
3 & - & - & Disc. & - & Tab. \ref{pretraining}.3 & 25.55 & 56.67 & 22.94 & 45.88 & 41.29&0.7\\
\hline
4 & \multirow{2}{*}{Input}  &Fbank& Fbank & Fbank&Tab. \ref{pretraining}.7 & 31.53 & 64.89&28.76 & 53.30& 48.39&1.9\\
5 & \multirow{2}{*}{Concat.} &Con.& Con. & Con.&Tab. \ref{pretraining}.8 & 27.66 & 57.16 & 25.26 & 46.88 & 43.04&7.3\\
6 &   &Disc.& Disc. & Disc.&Tab. \ref{pretraining}.9&29.28 & 60.40 & 26.67 & 49.61 & 45.02&1.3\\

\hline
7 &Cross-utterance   &Fbank& Fbank & Fbank & Tab. \ref{pretraining}.20 & 26.83 & $59.49^\dagger$ & 24.06 & 48.19 & $43.29^\dagger$&1.8\\ 
8 &Pooling Projection &Con.& Con. & Con. & Tab. \ref{pretraining}.23&23.53 & $52.26^\star$ & 21.23 & 42.27 & $38.04^\star$&7.0\\
9 & 32-dim & Disc.& Disc. & Disc. &Tab. \ref{pretraining}.24&24.91 &$55.26^\triangle$&22.37 &44.73 & 40.24&1.2\\ 

\hline
10 & Cross-utterance &Fbank& Fbank & Fbank &Tab. \ref{pretraining}.25&$26.51^\dagger$ & $58.88^\dagger$ & 23.83 & $47.63^\dagger$ & $42.86^\dagger$&1.9\\
11 & Encoder &Con.& Con. & Con. & Tab. \ref{pretraining}.28&$23.27^\star$ & $51.67^\star$ & 20.91 & $41.79^\star$ & $37.61^\star$&7.2\\
12 & embeddings &Disc.& Disc. & Disc. &Tab. \ref{pretraining}.29&$\textbf{24.69}^\triangle$ &$\textbf{54.77}^\triangle$&\textbf{22.16}&$\textbf{44.34}^\triangle$&$\textbf{39.88}^\triangle$&1.3\\

\bottomrule
\end{tabular}}
\vspace{-0.5cm}
\end{table}
\subsection{Performance on the DementiaBank Pitt Elderly Speech}
\vspace{-0.1cm}
In this section, we utilize the best pre-trained systems (Sys.1-3, 7-9, 20, 23, 24, 25, 28, 29 in Tab.\ref{pretraining}) for fine-tuning on elderly speech datasets. From Table~\ref{fine-tuning}, several trends can be found:

\textbf{1)} The best pre-trained systems for cross-utterance speech contexts fusion consistently exhibit the same performance trends when fine-tuning on elderly speech across all three input feature types (Fbank, continuous WavLM features, or WavLM discrete tokens). This consistency is observed in both the cross-utterance pooling projection approach (Sys.7-9) and the cross-utterance Encoder embeddings approach (Sys.10-12) when compared to the non-contextual systems that utilize the current utterance context only (Sys.1-3). In particular, the best fine-tuned discrete tokens based contextual Z-T system (Sys.12) outperforms the non-contextual baseline (Sys.3) by a statistically significant average WER reduction of \textbf{1.41}\% absolute (\textbf{3.4}\% relative) on the DementiaBank Pitt elderly speech.

\textbf{2)} The contextual Z-T systems using discrete token features as input (Sys.6,9,12) retain comparable WER reduction differences over the contextual Z-T system utilizing continuous WavLM features as input (Sys.5,8,11), while only requiring \textbf{17.1}\% to \textbf{18.0}\% of their training time.


%

\vspace{-0.3cm}
\section{Conclusion}
In this paper, we introduce discrete token-based cross-utterance speech contexts modelling for Z-T systems. Experiments on the 1000-hr GigaSpeech-M and DementiaBank Pitt elderly speech datasets show their superior efficacy and efficiency in modelling preceding, current, and future speech contexts utilizing Encoder embeddings concatenation or pooling projection of Encoder embeddings compared to contextual Z-T baselines that use Fbank or continuous WavLM features. The best-performing Z-T system utilizing discrete tokens outperforms the non-contextual baseline by statistically significant WER reductions of \textbf{0.39}\% and \textbf{1.41}\% absolute (\textbf{3.4}\% and \textbf{3.4}\% relative) on the two tasks, respectively. Model training time speedup ratios up to \textbf{4.36}x are obtained over continuous WavLM feature-based contextual Z-T systems, while retaining up to \textbf{98.0}\% of their WER reductions over non-contextual baselines. Future studies will explore SSL discrete token-based cross-utterance speech contexts for streaming Z-T. Using
WavLM discrete token features throughout for preceding, current, and future utterances’ contexts yielded the lowest WER of \textbf{11.15}\% and
\textbf{11.14}\%, achieving a SOTA performance. Future studies will explore SSL discrete token-based cross-utterance contexts for streaming Z-T ASR systems.
\vspace{-0.2cm}
\section{Acknowledgement}
This research is supported by Hong Kong RGC GRF grant No. 14200220, 14200021, 14200324, Innovation Technology Fund grant No. ITS/218/21 and the National Natural Science Foundation of China  (No. 62206171).
\bibliographystyle{IEEEtran}
\bibliography{mybib}

\begin{thebibliography}{10}
\providecommand{\url}[1]{#1}
\csname url@samestyle\endcsname
\providecommand{\newblock}{\relax}
\providecommand{\bibinfo}[2]{#2}
\providecommand{\BIBentrySTDinterwordspacing}{\spaceskip=0pt\relax}
\providecommand{\BIBentryALTinterwordstretchfactor}{4}
\providecommand{\BIBentryALTinterwordspacing}{\spaceskip=\fontdimen2\font plus
\BIBentryALTinterwordstretchfactor\fontdimen3\font minus \fontdimen4\font\relax}
\providecommand{\BIBforeignlanguage}[2]{{%
\expandafter\ifx\csname l@#1\endcsname\relax
\typeout{** WARNING: IEEEtran.bst: No hyphenation pattern has been}%
\typeout{** loaded for the language `#1'. Using the pattern for}%
\typeout{** the default language instead.}%
\else
\language=\csname l@#1\endcsname
\fi
#2}}
\providecommand{\BIBdecl}{\relax}
\BIBdecl

\bibitem{wei2022leveraging}
K.~Wei, Y.~Zhang, S.~Sun, L.~Xie, and L.~Ma, ``Leveraging acoustic contextual representation by audio-textual cross-modal learning for conversational asr,'' \emph{INTERSPEECH}, 2022.

\bibitem{kim2018dialog}
S.~Kim and F.~Metze, ``{Dialog-context aware end-to-end speech recognition},'' in \emph{SLT Workshop}, 2018.

\bibitem{hou2022bring}
J.~Hou \emph{et~al.}, ``{Bring dialogue-context into RNN-T for streaming ASR},'' \emph{INTERSPEECH}, 2022.

\bibitem{chang2023context}
S.-Y. Chang \emph{et~al.}, ``{Context-aware end-to-end ASR using self-attentive embedding and tensor fusion},'' in \emph{IEEE ICASSP}, 2023.

\bibitem{li2022recent}
J.~Li \emph{et~al.}, ``{Recent advances in end-to-end automatic speech recognition},'' \emph{APSIPA Transactions on Signal and Information Processing}, 2022.

\bibitem{chen2021developing}
X.~Chen, Y.~Wu, Z.~Wang, S.~Liu, and J.~Li, ``Developing real-time streaming transformer transducer for speech recognition on large-scale dataset,'' in \emph{ICASSP}, 2021.

\bibitem{irie2019training}
K.~Irie \emph{et~al.}, ``{Training language models for long-span cross-sentence evaluation},'' in \emph{ASRU Workshop}, 2019.

\bibitem{xiong2018session}
W.~Xiong, L.~Wu, J.~Zhang, and A.~Stolcke, ``Session-level language modeling for conversational speech,'' in \emph{EMNLP}, 2018.

\bibitem{dai2019transformer}
Z.~Dai \emph{et~al.}, ``{Transformer-xl: attentive language models beyond a fixed-length context},'' \emph{ACL}, 2019.

\bibitem{liu2020contextualizing}
D.-R. Liu \emph{et~al.}, ``{Contextualizing ASR lattice rescoring with hybrid pointer network language model},'' \emph{INTERSPEECH}, 2020.

\bibitem{liu2013use}
X.~Liu, M.~J.~F. Gales, and P.~C. Woodland, ``Use of contexts in language model interpolation and adaptation,'' \emph{CSL}, 2013.

\bibitem{beltagy2020longformer}
I.~Beltagy \emph{et~al.}, ``{Longformer: the long-document transformer},'' \emph{arXiv preprint arXiv:2004.05150}, 2020.

\bibitem{sun2021transformer}
G.~Sun \emph{et~al.}, ``{Transformer language models with LSTM-based cross-utterance information representation},'' in \emph{IEEE ICASSP}, 2021.

\bibitem{chang2021context}
F.-J. Chang \emph{et~al.}, ``{Context-aware transformer transducer for speech recognition},'' in \emph{ASRU Workshop}, 2021.

\bibitem{chen2022factorized}
X.~Chen, Z.~Meng \emph{et~al.}, ``Factorized neural transducer for efficient language model adaptation,'' in \emph{IEEE ICASSP}, 2022.

\bibitem{gong2023longfnt}
X.~Gong \emph{et~al.}, ``Longfnt: long-form speech recognition with factorized neural transducer,'' in \emph{IEEE ICASSP}, 2023.

\bibitem{gong2024advanced}
X.~Gong, Y.~Wu, J.~Li, S.~Liu, R.~Zhao, X.~Chen, and Y.~Qian, ``Advanced long-content speech recognition with factorized neural transducer,'' \emph{IEEE/ACM TASLP}, 2024.

\bibitem{wang2025phonepurityguideddiscretetokens}
\BIBentryALTinterwordspacing
H.~Wang, X.~Xie, M.~Geng, S.~Hu, H.~Xu, Y.~Chen, Z.~Li, J.~Deng, and X.~Liu, ``Phone-purity guided discrete tokens for dysarthric speech recognition,'' 2025. [Online]. Available: \url{https://arxiv.org/abs/2501.04379}
\BIBentrySTDinterwordspacing

\bibitem{baevski2019effectiveness}
A.~Baevski, M.~Auli, and A.~Mohamed, ``Effectiveness of self-supervised pre-training for speech recognition,'' \emph{arXiv preprint arXiv:1911.03912}, 2019.

\bibitem{chang2023exploration}
X.~Chang \emph{et~al.}, ``Exploration of efficient end-to-end asr using discretized input from self-supervised learning,'' \emph{arXiv preprint arXiv:2305.18108}, 2023.

\bibitem{yang2023towards}
Y.~Yang \emph{et~al.}, ``Towards universal speech discrete tokens: a case study for asr and tts,'' \emph{arXiv preprint arXiv:2309.07377}, 2023.

\bibitem{guo2025recent}
Y.~Guo, Z.~Li, H.~Wang, B.~Li, C.~Shao, H.~Zhang, C.~Du, X.~Chen, S.~Liu, and K.~Yu, ``Recent advances in discrete speech tokens: A review,'' \emph{arXiv preprint arXiv:2502.06490}, 2025.

\bibitem{wang2024evaluating}
S.~Wang and {\'E}.~Sz{\'e}kely, ``Evaluating text-to-speech synthesis from a large discrete token-based speech language model,'' \emph{arXiv preprint arXiv:2405.09768}, 2024.

\bibitem{shen2024acoustic}
F.~Shen, Y.~Guo, C.~Du, X.~Chen, and K.~Yu, ``Acoustic bpe for speech generation with discrete tokens,'' in \emph{ICASSP 2024-2024 IEEE International Conference on Acoustics, Speech and Signal Processing (ICASSP)}.\hskip 1em plus 0.5em minus 0.4em\relax IEEE, 2024, pp. 11\,746--11\,750.

\bibitem{lee2024high}
J.~Y. Lee, M.~Jeong, M.~Kim, J.-H. Lee, H.-Y. Cho, and N.~S. Kim, ``High fidelity text-to-speech via discrete tokens using token transducer and group masked language model,'' \emph{arXiv preprint arXiv:2406.17310}, 2024.

\bibitem{chang2024interspeech2024challengespeech}
\BIBentryALTinterwordspacing
X.~Chang, J.~Shi, J.~Tian, Y.~Wu, Y.~Tang, Y.~Wu, S.~Watanabe, Y.~Adi, X.~Chen, and Q.~Jin, ``The interspeech 2024 challenge on speech processing using discrete units,'' 2024. [Online]. Available: \url{https://arxiv.org/abs/2406.07725}
\BIBentrySTDinterwordspacing

\bibitem{chang2023exploringspeechrecognitiontranslation}
\BIBentryALTinterwordspacing
X.~Chang, B.~Yan, K.~Choi, J.~Jung, Y.~Lu, S.~Maiti, R.~Sharma, J.~Shi, J.~Tian, S.~Watanabe, Y.~Fujita, T.~Maekaku, P.~Guo, Y.-F. Cheng, P.~Denisov, K.~Saijo, and H.-H. Wang, ``Exploring speech recognition, translation, and understanding with discrete speech units: A comparative study,'' 2023. [Online]. Available: \url{https://arxiv.org/abs/2309.15800}
\BIBentrySTDinterwordspacing

\bibitem{chen2024loss}
Q.~Chen, W.~Wang, Q.~Zhang, S.~Zheng, S.~Zhang, C.~Deng, Y.~Ma, H.~Yu, J.~Liu, and C.~Zhang, ``Loss masking is not needed in decoder-only transformer for discrete-token-based asr,'' in \emph{ICASSP}.\hskip 1em plus 0.5em minus 0.4em\relax IEEE, 2024, pp. 11\,056--11\,060.

\bibitem{chen2022wavlm}
S.~Chen \emph{et~al.}, ``Wavlm: large-scale self-supervised pre-training for full stack speech processing,'' \emph{IEEE JSTSP}, 2022.

\bibitem{graves2012sequence}
A.~Graves, ``Sequence transduction with recurrent neural networks,'' \emph{arXiv preprint arXiv:1211.3711}, 2012.

\bibitem{ghodsi2020rnn}
M.~Ghodsi, X.~Liu \emph{et~al.}, ``Rnn-transducer with stateless prediction network,'' in \emph{IEEE ICASSP}, 2020.

\bibitem{cui23_interspeech}
M.~Cui \emph{et~al.}, ``{Towards effective and compact contextual representation for conformer transducer speech recognition systems},'' in \emph{INTERSPEECH}, 2023.

\bibitem{chen2021gigaspeech}
G.~Chen, ``Gigaspeech: An evolving, multi-domain asr corpus with 10,000 hours of transcribed audio,'' \emph{arXiv preprint arXiv:2106.06909}, 2021.

\bibitem{becker1994natural}
J.~T. Becker, F.~Boiler, O.~L. Lopez, J.~Saxton, and K.~L. McGonigle, ``{The natural history of Alzheimer's disease: description of study cohort and accuracy of diagnosis},'' \emph{Archives of neurology}, 1994.

\bibitem{ye2021development}
Z.~Ye, S.~Hu, J.~Li, X.~Xie, M.~Geng, J.~Yu, J.~Xu, B.~Xue, S.~Liu, X.~Liu \emph{et~al.}, ``{Development of the CUHK Elderly speech recognition system for neurocognitive disorder detection using the dementiabank corpus},'' in \emph{ICASSP}, 2021.

\bibitem{yao2023zipformer}
Z.~Yao, L.~Guo \emph{et~al.}, ``Zipformer: A faster and better encoder for automatic speech recognition,'' \emph{arXiv preprint arXiv:2310.11230}, 2023.

\bibitem{pasad2023comparativelayerwiseanalysisselfsupervised}
\BIBentryALTinterwordspacing
A.~Pasad, B.~Shi, and K.~Livescu, ``Comparative layer-wise analysis of self-supervised speech models,'' 2023. [Online]. Available: \url{https://arxiv.org/abs/2211.03929}
\BIBentrySTDinterwordspacing

\bibitem{pallet1990tools}
D.~S. Pallet \emph{et~al.}, ``{Tools for the analysis of benchmark speech recognition tests},'' in \emph{ICASSP}, 1990.

\bibitem{gillick1989some}
L.~Gillick and S.~J. Cox, ``{Some statistical issues in the comparison of speech recognition algorithms},'' in \emph{ICASSP}, 1989.

\bibitem{hori2020transformer}
T.~Hori, N.~Moritz, C.~Hori, and J.~Le~Roux, ``Transformer-based long-context end-to-end speech recognition.'' in \emph{INTERSPEECH}, 2020.

\bibitem{hori2021advanced}
T.~Hori \emph{et~al.}, ``{Advanced Long-context E2E Speech Recognition using Context-expanded Transformers},'' \emph{INTERSPEECH}, 2021.

\bibitem{fox2024updated}
J.~D. Fox, D.~Raj, N.~Delworth, Q.~McNamara, C.~Miller, and M.~Jett{\'e}, ``Updated corpora and benchmarks for long-form speech recognition,'' in \emph{ICASSP 2024-2024 IEEE International Conference on Acoustics, Speech and Signal Processing (ICASSP)}, 2024.

\end{thebibliography}

\end{document}